\documentclass[10pt,twocolumn,letterpaper]{article}

\usepackage{cvpr}              

\usepackage{graphicx}
\usepackage{amsmath}
\usepackage{amssymb}
\usepackage{booktabs}
\usepackage{multirow}
\usepackage{bm}
\usepackage{threeparttable}
\newcommand{\tabincell}[2]{\begin{tabular}{@{}#1@{}}#2\end{tabular}}

\usepackage[accsupp]{axessibility}

\usepackage[pagebackref,breaklinks,colorlinks]{hyperref}

\usepackage[capitalize]{cleveref}
\crefname{section}{Sec.}{Secs.}
\Crefname{section}{Section}{Sections}
\Crefname{table}{Table}{Tables}
\crefname{table}{Tab.}{Tabs.}

\def\cvprPaperID{4143} 
\def\confName{CVPR}
\def\confYear{2022}

\begin{document}

\title{Shapley-NAS: Discovering Operation Contribution for Neural \\Architecture Search}

\author{Han Xiao\textsuperscript{1,2},
	Ziwei Wang\textsuperscript{1,2},
	Zheng Zhu\textsuperscript{1,2},
	Jie Zhou\textsuperscript{1,2},
	Jiwen Lu\textsuperscript{1,2}\thanks{Corresponding author},\\	
	\textsuperscript{1}Department of Automation, Tsinghua University, China\\
	\textsuperscript{2}Beijing National Research Center for Information Science and Technology, China\\
	{\tt \small \{h-xiao20,wang-zw18\}@mails.tsinghua.edu.cn; zhengzhu@ieee.org; \{jzhou,lujiwen\}@tsinghua.edu.cn}
}
\maketitle

\begin{abstract}
In this paper, we propose a Shapley value based method to evaluate operation contribution (Shapley-NAS) for neural architecture search. Differentiable architecture search (DARTS) acquires the optimal architectures by optimizing the architecture parameters with gradient descent, which significantly reduces the search cost. However, the magnitude of architecture parameters updated by gradient descent fails to reveal the actual operation importance to the task performance and therefore harms the effectiveness of obtained architectures. By contrast, we propose to evaluate the direct influence of operations on validation accuracy. To deal with the complex relationships between supernet components, we leverage Shapley value to quantify their marginal contributions by considering all possible combinations. Specifically, we iteratively optimize the supernet weights and update the architecture parameters by evaluating operation contributions via Shapley value, so that the optimal architectures are derived by selecting the operations that contribute significantly to the tasks. Since the exact computation of Shapley value is NP-hard, the Monte-Carlo sampling based algorithm with early truncation is employed for efficient approximation, and the momentum update mechanism is adopted to alleviate fluctuation of the sampling process. Extensive experiments on various datasets and various search spaces show that our Shapley-NAS outperforms the state-of-the-art methods by a considerable margin with light search cost. The code is available at \href{https://github.com/Euphoria16/Shapley-NAS.git}{https://github.com/Euphoria16/Shapley-NAS.git.}
\end{abstract}

\section{Introduction}

Neural architecture search (NAS) has attracted great interest in deep learning since it discovers the optimal architecture from a large search space of network components according to task performance and hardware configurations. Pioneering works applied reinforcement learning \cite{zoph2016neural}, evolutionary algorithms \cite{real2019regularized, wang2020apq}, and Bayesian optimization \cite{liu2018progressive} for the architecture search, but the large computational overhead prohibits practical deployment of NAS algorithms. Therefore, it is desirable to design highly efficient search strategies without performance degradation.

To reduce the search cost of architecture search, several efficient search strategies have been presented including one-shot NAS \cite{pham2018efficient}, network transformation \cite{cai2018efficient}, and architecture optimization \cite{luo2018neural}. Among these approaches, one-shot NAS preserves the optimal sub-networks from the over-parameterized supernet with weight sharing, which prevents the time-consuming exhaustive training for model evaluation. In particular, DARTS \cite{liu2018darts} converted the discrete operation selection into continuous mixing weights learning and iteratively optimized the architecture parameters and supernet weights by gradient descent with significantly reduced search cost. However, the magnitude of architecture parameters in DARTS cannot reflect the actual operation importance in general \cite{wang2021rethinking, zhou2021exploiting, yu2019evaluating}. That is, the operation with the largest parameter magnitude does not necessarily result in the highest validation accuracy, which degrades the performance of derived architectures. 
\begin{figure*}[t]
    \begin{center}
    \includegraphics[width=0.9 \linewidth]{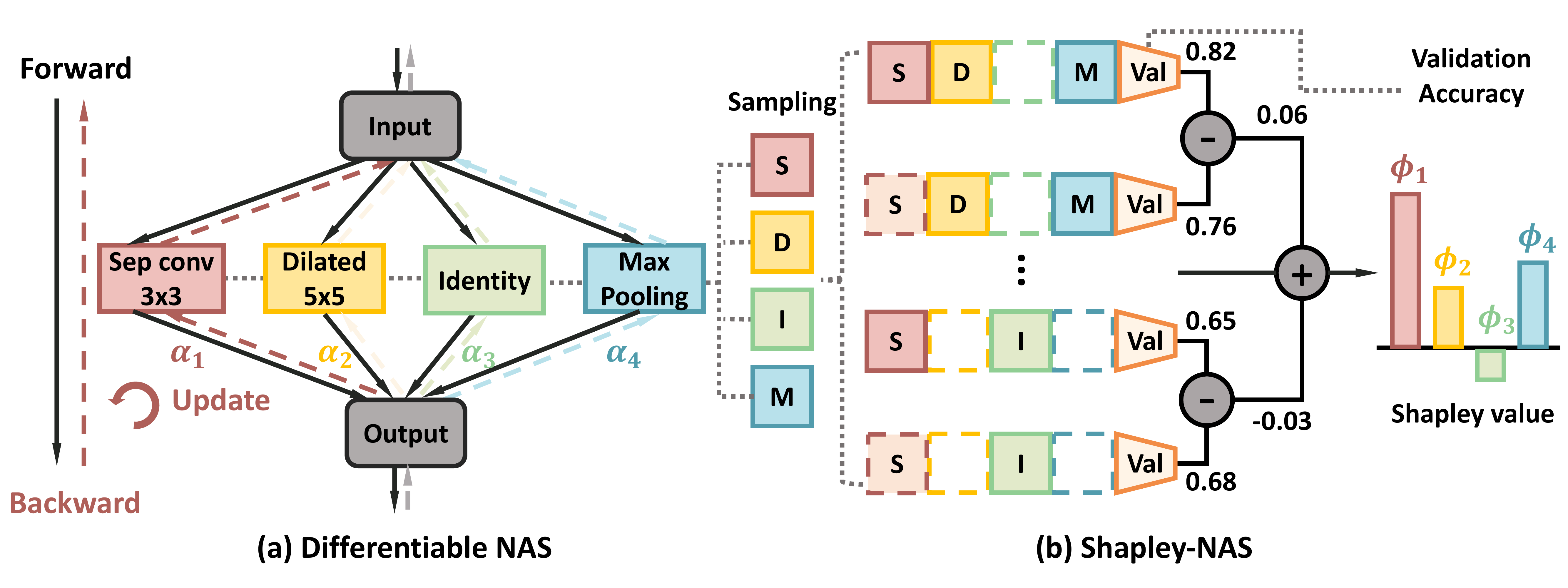}
    \vspace{-1em}
    \caption{The comparison between DARTS and our Shapley-NAS. (a) DARTS constructs a weight-sharing supernet that consists of all candidate operations. The architecture parameters are optimized by gradient descent, which fails to reflect the importance of operations \cite{wang2021rethinking, zhou2021exploiting, yu2019evaluating}. (b) The proposed Shapley-NAS method directly evaluates the marginal contribution of operations to the task performance, according to the validation accuracy difference of all possible operation subsets and their counterparts without the given operation.}
\vspace{-2.3em}
    \label{fig:main}
\end{center}
\end{figure*}

In this paper, we present a Shapley-NAS method to evaluate the operation contribution via the Shapley value of supernet components for neural architecture search. Instead of relying on the magnitude of architecture parameters updated by gradient descent, we consider their practical influences on task performance and propose to directly evaluate their contributions to the validation accuracy. Moreover, we observe that the operations in the supernet are related to each other: combinations of operations might have different joint influences on performance compared with their separate ones. In order to deal with such complex relationships, we leverage Shapley value \cite{shapley1953value, roth1988shapley}, an important solution to attribute contributions to players in cooperative game theory. \cref{fig:main} shows the differences between our Shapley-NAS and existing DARTS methods. Shapley value directly measures the contributions of operations according to the validation accuracy difference. Meanwhile, it considers all possible combinations and quantifies the average marginal contribution to handle complex relationships between individual elements. Benefiting from these, Shapley value is effective for obtaining operation importance that is highly correlated with task performance. Since computing the exact Shapley value is NP-hard, we employ the Monte-Carlo sampling with early truncation for operation permutation set sampling to approximate it efficiently. Finally, we optimize the supernet weights and update the architecture parameters iteratively, where the momentum update mechanism is adopted to alleviate the fluctuation caused by the sampling process. We empirically demonstrate that the obtained Shapley value has a higher correlation with task performance compared with DARTS. We conducted extensive experiments on different datasets across various search spaces, where our Shapley-NAS outperforms the state-of-the-art architecture search methods. We achieve an error rate of $2.43\%$ on CIFAR-10 \cite{krizhevsky2009learning} on the search space of DARTS and obtain the top-1 accuracy of $23.9\%$ on ImageNet \cite{deng2009imagenet} under the mobile setting. Furthermore, our Shapley-NAS acquires the optimal architectures on CIFAR-10 and CIFAR-100 and near-optimal solutions on ImageNet-16-120 of the NAS-Bench-201 benchmark \cite{dong2020bench}.

\section{Related Work}
\textbf{Differentiable NAS: }Differentiable architecture search (DARTS) was first proposed by Liu \etal \cite{liu2018darts} with the goal of reducing the heavy search cost in NAS. They apply a continuous relaxation to the graphical architecture representation, thus enabling efficient gradient descent to solve the bi-level optimization objective for architecture search. PC-DARTS \cite{xu2019pc} further proposed to only search the partially-connected operations by leveraging the redundancy in network space to further reduce the memory overhead. Despite the computation efficiency of DARTS, several works have challenged its generalizability \cite{chen2019progressive, li2020sgas, xie2018snas, yu2019evaluating} and stability \cite{chen2020drnas, chen2020stabilizing, zhang2021idarts, zela2019understanding, wang2021rethinking}. In order to reduce the bias of DARTS for operation selection, SNAS \cite{xie2018snas} and GDAS \cite{dong2019searching} introduced stochasticity into the supernet training and adopted the differentiable Gumbel-Softmax trick\cite{jang2016categorical} for gradient estimation. SGAS \cite{li2020sgas} greedily chose and pruned the candidate operations based on edge importance, selection certainty, and selection stability to alleviate the performance gap between the search and evaluation phase. RobustDARTS \cite{zela2019understanding} found that the stability of DARTS is highly correlated with dominant eigenvalue of the Hessian of validation loss with respect to the architecture parameters. Therefore, they performed early stop regularization according to the largest eigenvalue to avoid poor generation. SmoothDARTS \cite{chen2020stabilizing} further smoothed the loss landscape via perturbation-based regularization. However, recent studies \cite{wang2021rethinking, zhou2021exploiting, yu2019evaluating} have demonstrated the magnitude of architecture parameters in the DARTS framework fails to reveal the actual operation importance, which greatly degrades the performance of architectures derived from the search phase.

\textbf{Shapley value: }Shapley value has been well studied in the cooperative game theory as a fair contribution distribution method \cite{shapley1953value, roth1988shapley}. Recently, Shapley value has been adopted in explainable machine learning to discover the importance of different elements, which can be divided into three groups: explaining feature importance \cite{mase2019explaining, lundberg2017unified, lundberg2020local, ancona2019explaining, strumbelj2010efficient}, model component importance \cite{ancona2020shapley, wang2021shapley, ghorbani2020neuron}, and data importance \cite{jia2019towards, yona2021s}. For the first regard, Ancona \etal \cite{ancona2019explaining} conducted an axiomatic comparison to show the advantage of the Shapley value over the attribution methods for feature map explanation in deep networks. SHAP \cite{lundberg2017unified} presented the additive feature attribution based on the Shapley value of features to acquire higher consistency with human intuition. For model component importance explanation, ShapNets \cite{wang2021shapley} leveraged the Shapley transform that transforms the input into Shapley representations so that the network prediction can be explained during the forward pass. Neuron Shapley \cite{ghorbani2020neuron} identified the most important filters in neural networks and demonstrated potential applications to improve the accuracy, fairness, and robustness of the model prediction. Moreover, Ghorbani \etal \cite{ghorbani2019data} quantified the contribution of individual data points which effectively identified the outliers and corrupted data. Since computing the exact Shapley value is NP-hard, Monte-Carlo sampling \cite{ghorbani2019data, ghorbani2020neuron}, perturbation-based approximation \cite{ancona2019explaining}, influence function, and many others were presented for efficient estimation of Shapley value. In this paper, we extend the Shapley value to operation importance evaluation in the DARTS framework, so that the optimal architectures are derived by selecting the operations that contribute significantly to the performance.

\section{Approach}
\label{headings}

In this section, we first briefly introduce differentiable architecture search (DARTS), which suffers from degenerate architectures due to the mismatch between the architecture parameters and operation importance. Then we propose to directly evaluate the influence of operations on the task performance and introduce Shapley value to quantify their relative contributions at the presence of complex relationships between different operations. We also present the Monte-Carlo sampling algorithm with early truncation for efficient approximation of Shapley value. Finally, we propose Shapley-based architecture search (Shapley-NAS) which can effectively identify the optimal architectures with the most important operations from the large search space.

\subsection{Preliminaries}

The differentiable architecture search (DARTS) is one of the most popular solutions to identify effective architectures, as it largely reduces the search cost by relaxing the architecture search to continuous mixture weights learning. Following prior works \cite{zoph2018learning, liu2018progressive, real2019regularized}, DARTS searches for the best cell structure and constructs the supernet by repetitions of normal and reduction cells. Each cell is represented by a directed acyclic graph (DAG) with $\mathcal{N}$ nodes and $\mathcal{E}$ edges, where each node $x^{(i)}$ defines a latent representation and each edge $(i,j)$ is associated with an operation $o^{(i,j)}$. The core idea of DARTS is to apply continuous relaxation to the search space to perform the gradient-based search. Concretely, the intermediate node is computed as a softmax mixture of candidate operations:
\begin{equation}
    \bar{o}^{(i,j)}(x^{(i)})=\sum_{o\in \mathcal{O}} \frac{\text{exp}(\alpha_{o}^{(i,j)})}
    {\sum_{o^{'}\in \mathcal{O}} \text{exp}(\alpha_{o^{'}}^{(i,j)})} o(x^{(i)}),
\end{equation}
where $\mathcal{O}$ is the set of all candidate operations and $\alpha_{o}^{(i,j)}$ denotes the mixing weight of operation $o^{(i,j)}$ to construct the supernet. With such relaxation, the architecture search can be performed by jointly optimizing the network weight $w$ and architecture parameters $\alpha$ in a differentiable manner with the following bi-level objective:
\begin{equation}
	       \begin{aligned}
        	\label{bi_level}
            \min_{\alpha}\  \ \mathcal{L}_{val} (w^*, \alpha)  \ \  \ \ \
            \text{ s.t.}\  \ \ \ \ w^* = \arg\min_{w}\ \mathcal{L}_{train}(w,\alpha).
        \end{aligned}       
\end{equation}

During the search stage, the weight-sharing supernet containing all these candidate operations is optimized by gradient descent. At the end of the search stage, the final architecture is derived by selecting the operation with the largest architecture parameter $\alpha$ on every edge across all operation choices, $o^{(i,j)}=\arg \max _{o\in\mathcal{O}} \alpha_{o}^{(i,j)}$. 

\subsection{Operation Importance Evaluation}
The magnitude-based architecture selection process in DARTS relies on an important assumption that the magnitude of architecture parameters represents the operation importance. In other words, it supposes that operations with low magnitude of $\alpha$ result in weak feature representations and thus have little contribution to the network performance. However, recent studies \cite{wang2021rethinking, zhou2021exploiting, yu2019evaluating} have shown that the value of architecture parameters does not necessarily reflect the actual operation contribution. In many cases, the operation with the largest $\alpha$ does not result in the highest validation accuracy. Therefore, selecting the best operations based on values of $\alpha$ may lead to significant performance degradation at the evaluation phase. 

\begin{figure}
   \centering
  \begin{subfigure}{0.37\linewidth}\vspace{0.2em}
  \includegraphics[ width=1.0\textwidth]{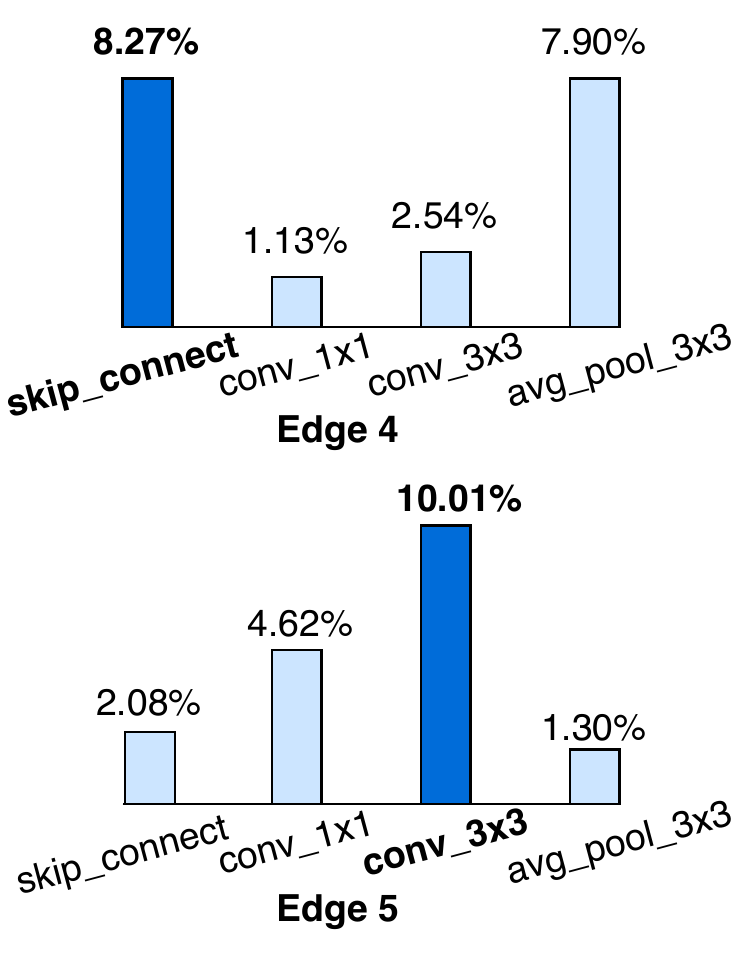}
    \caption{{Removing one operation on each edge}}
    \label{fig:2-1}
  \end{subfigure}
   \hfill 
  \begin{subfigure}{0.59\linewidth}
    \includegraphics[ width=1.0\textwidth]{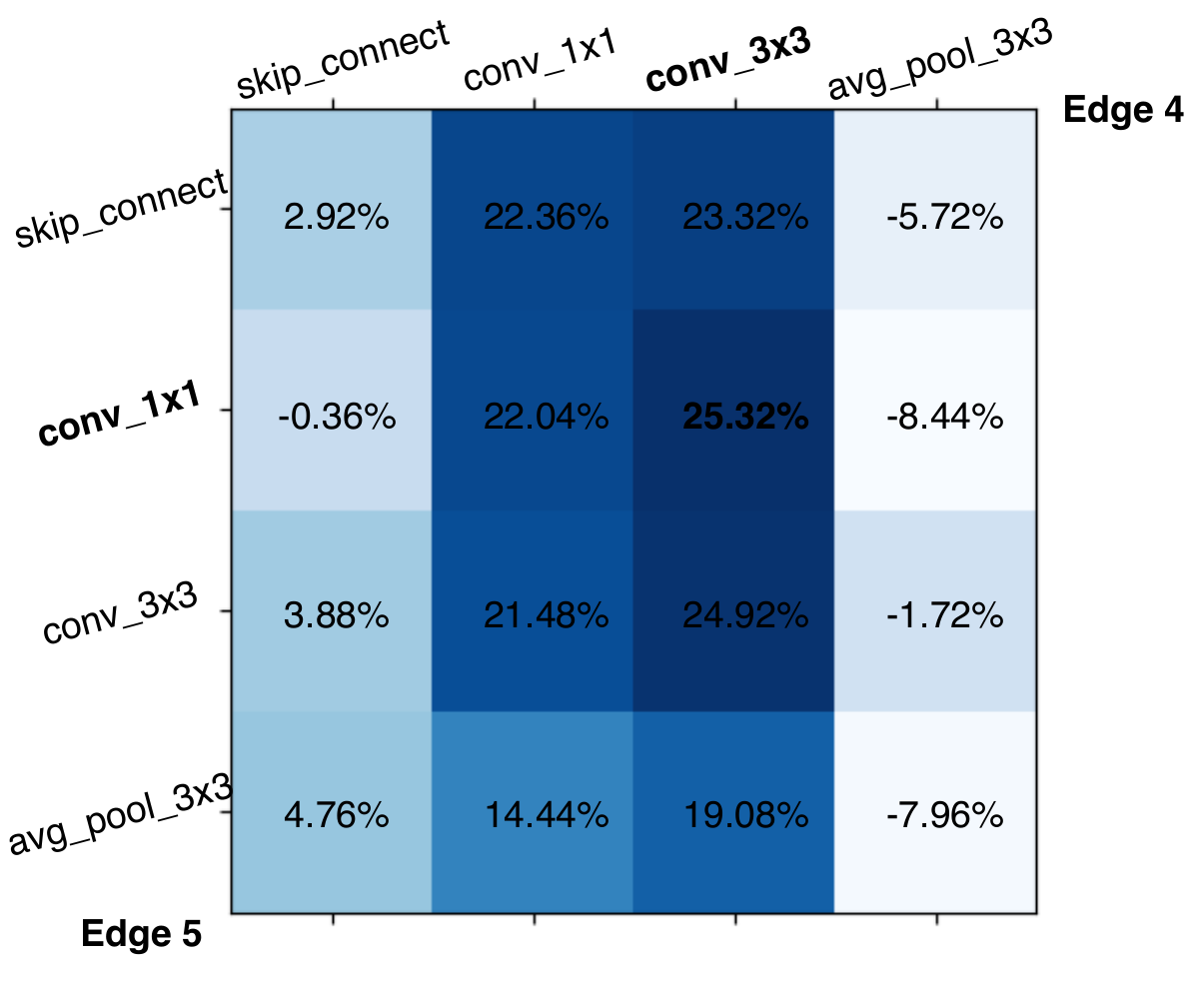}
    \caption{{Removing one operation on both edges at the same time}}
    \label{fig:2-2}
  \end{subfigure}
    \vspace{-0.5em}
    \caption{\small{ The performance drop caused by (a) removing the target operation on the 4th and 5th edge separately (b) removing one operation on both edges at the same time and enumerating all combinations.}}
  \label{fig:2}
  \vspace{-1.8em}
\end{figure}

To solve this problem, we propose to perform the architecture search by identifying operations that contribute the most to the validation accuracy. \cite{wang2021rethinking} performs a similar evaluation of operation contribution by removing the target operation from the supernet while keeping all other operations to obtain the performance drop. However, we observe that the operations in the supernet are not independent of each other. To demonstrate the underlying relationships between operations on different edges, we remove only one operation separately on the 4th edge and the 5th edge from the supernet pretrained on the NAS-Bench-201 space, and re-evaluate the supernet accuracy. As shown in \cref{fig:2-1}, removing the \emph{skip\_connect} operation on the 4th edge and the \emph{conv\_3x3} on the 5th edge leads to the most dramatic performance drop. However, we find the impacts of combinations of the two edges differ from the simple accumulation of their separate influence. We remove one operation for both edges simultaneously and enumerate all candidate operation combinations to show the results. As \cref{fig:2-2} illustrates, removing \emph{conv\_3x3} on the 4th edge and \emph{conv\_1x1} on the 5th edge results in the most significant degradation, while removing the combination of \emph{skip\_connect} and \emph{conv\_3x3} only lead to $3.88\%$ performance drop. 

This observation reveals the complex relationships between different operations on different edges: some operations can collaborate with each other and thus have a significant joint contribution to the supernet performance. To deal with such relationships, we leverage Shapley value \cite{shapley1953value, roth1988shapley}, an important solution from cooperative game theory, to evaluate the individual contribution. Specifically, the differentiable architecture search process can be uniquely mapped into a cooperative game, providing a practical scheme to quantify the operation contribution based on Shapley value. In a cooperative game, $N$ players associate with each other, and a value function $V$ maps each subset of players $S\subseteq N$ to a real value $V(S)$, which represents the expected payoff the players can obtain by cooperation. In differentiable NAS, the supernet is composed of several layers with identical cell structures, and each cell has $|\mathcal{E}|$ edges each with $|\mathcal{O}|$ operations. Therefore, a set of individual operations, $N=\mathcal{O}\times \mathcal{E}=\{ {o^{(i,j)}} \}_{o\in \mathcal{O},(i,j) \in \mathcal{E}}$, can be modeled as players in the cooperative game, where all players work together towards the supernet's performance $V(N)$. Shapley value is utilized to distribute the total performance gains $V(N)$ to each player in $N$. In our problem, for operation $o^{(i,j)}$, its Shapley value $\phi_o^{(i,j)}$ can be computed as:
\begin{equation}
	\phi_o^{(i,j)}(V)=\frac{1}{|N|}{\sum_{S\subseteq N\setminus\{o^{(i,j)}\}} \frac{V(S\cup \{o^{(i,j)}\})-V(S)}{\binom{|N|-1}{|S|}}}
	\label{shapley_compute}
\end{equation}

The Shapley value represents the average marginal contribution of the operation to the network performance, which is obtained by evaluating the performance difference between all operation subsets and their counterparts without the given operation. Here we use the validation accuracy as the value function $V$ to measure the network performance. It has been proved that the formulation of Shapley value in \cref{shapley_compute} makes it the only method to quantify individual contributions that uniquely satisfies the following properties\cite{shapley1953value}, which we interpret according to our problem:

\textbf{Efficiency} \quad The performance of the entire supernet is the sum of contributions of individual operations, i.e. $\sum_{o^{(i,j)} \in N}\phi_o^{(i,j)}=V(N)$.

\textbf{Null Player}\quad If the operation has no impact on the performance when added to or removed from any subsets of the supernet, then its contribution is zero. That is, if $V(S)=V(S\cup\{o^{(i,j)}\})$ for any operation subset $S\subseteq N\setminus\{o^{(i,j)}\}$, we can derive $ \phi_o^{(i,j)}=0$. For example, the zero operation in the search space of DARTS has no impact on the final performance and thus has zero contribution.

\textbf{Symmetry}\quad If two different operations could be exchanged without affecting the performance, they should be assigned with equal contributions. For any operation subset $S\subseteq N\setminus\{o^{(i,j)}, o'^{(k,l)}\}$, $V(S\cup \{ o^{(i,j)} \}) =	V(S\cup \{ {o'^{(k,l)}} \})$, then we have $\phi_{o}^{(i,j)}=\phi_{o'}^{(k,l)}$.

\subsection{Shapley Value Approximation}
\label{estimation_shapley}
 Although Shapley value can be considered as a desirable attribution metric for quantifying the contribution of operations, directly computing Shapley value from \cref{shapley_compute} requires $2^{|\mathcal{O}|\times |\mathcal{E}|}$ network evaluations caused by enumerating all possible subsets. Therefore, the exact computation of Shapley value becomes expensive since $|\mathcal{O}|\times |\mathcal{E}|$ in the common search space is usually large.
 To efficiently estimate the Shapley value, we present an approximate method based on Monte-Carlo sampling \cite{castro2009polynomial}. Specifically, 
 the Shapley value of operation $o^{(i,j)}$(denoted as $o$ for simplicity) is equivalent to estimating the mean of a random variable, which can be written as:
\begin{equation}
		\phi_o(V)=\frac{1}{N!} \sum_{R\in \pi(N)}[V(Pre_{o}(R)\cup \{o\})-V(Pre_{o}(R))]
		\label{monte}
\end{equation}
where $\pi(N)$ denotes the set of permutations of all elements in $N$, and $Pre_{o}(R)$ is the set of predecessors of $o$ in a given permutation $R\in \pi(N)$. Based on \cref{monte}, we can get an unbiased approximation of every operation's Shapley value by sampling permutations of operation set $N$. Notably, the Monte-Carlo estimation reduces the exponential calculation complexity to polynomial-time $M\times (|\mathcal{O}|\times |\mathcal{E}|)$, where $M$ is the number of samples. Although this sampling-based estimation requires repetitions of accuracy evaluation on the validation set, it only includes the forward process through the supernet with no need for back-propagation, thus enabling efficient approximation of Shapley value.

Moreover, we find when the number of operations in $Pre_{o}(R)$ becomes too small, the task performance degrades dramatically and yields unstable sampling results. Therefore, to reduce the fluctuation of Shapley value estimation, we utilize the early truncation technique during the Monte-Carlo sampling procedure. Specifically, when the masked out operations lead to an extreme performance drop exceeding a pre-defined threshold $\eta$, we break off the current sampling. This early truncation technique also reduces nearly half of computation cost, which makes the overall computational overheads comparable with gradient-based architecture parameter optimization in DARTS.

 \subsection{Shapley-based Architecture Search}     
 We leverage the Shapley value of operations to guide the architecture search to find the best solutions as it reveals the actual operation contribution to performance. \cref{fig:main} shows the difference between our Shapley-NAS and conventional differential NAS. Instead of updating the architecture parameters by gradient descent, we utilize the Shapley value to represent the relative strength of operations. Specifically, the search objective should be modified as follows:
\begin{equation}
	       \begin{aligned}
        	\label{eq:shapley-nas}
            \alpha \propto \phi(\mathcal{L}_{val}(w^*,\alpha)))  \ \  \ \ \
            \text{ s.t.}\  \ \ \ \ w^* = \arg\min_{w}\ \mathcal{L}_{train}(w,\alpha).
        \end{aligned}       
\end{equation}
 
 Since solving the above problem exactly is impractical, we optimize this objective via an approximate way. We update $\bm{\alpha}$ according to the Shapley value estimated by the algorithm presented in \cref{estimation_shapley}:
\begin{equation}
\bm{\alpha}_{t}=\bm{\alpha}_{t-1} + \epsilon \cdot \frac{\bm{s}_t}{||\bm{s}_t||_2} 	
\end{equation}
 where $\bm{\alpha}_{t}$ means the architecture parameter at the $t$-th step during the optimization, $\bm{s}_t$ represents the accumulated Shapley value in the $t_{th}$ step, $||\cdot||_2$ is the $L_2$ norm and $\epsilon$ is the step size. 
 We iteratively optimize $\bm{w}_t$ by descending $\nabla L_{train}(\bm{w}_{t-1}, \bm{\alpha}_{t-1})$ and update architecture parameters $\bm{\alpha}$ until convergence.
 To reduce undesired fluctuation in updating caused by random sampling, we introduce the momentum into the iteration to stabilize the optimization:
 \begin{equation}
\bm{s}_{t}=\mu \cdot \bm{s}_{t-1} + (1-\mu) \cdot \frac{\bm{\phi}({Acc}_{val}(\bm{w}_{t-1},\bm{\alpha}_{t-1}))}{||\bm{\phi}({Acc}_{val}(\bm{w}_{t-1},\bm{\alpha}_{t-1}))||_2} 	
\end{equation}
where $\mu$ is the momentum coefficient that balances the accumulated Shapley value and the current sampling result, $Acc_{val}$ is the validation accuracy used as value function, and $\bm{w}_{t-1}$ is the supernet weights at ${(t-1)}$ step. After the search stage, we derive the final architecture by selecting the operation with the largest contribution on each edge.

\section{Experiments}
In this part, we conducted extensive experiments to evaluate our method on the DARTS search space with CIFAR-10 \cite{krizhevsky2009learning} and ImageNet \cite{deng2009imagenet} for image classification, as well as on a widely used NAS benchmark dataset, NAS-Bench-201 \cite{dong2020bench}. We first introduce the datasets and implementation details of our Shapley-NAS. In the following ablation study, we analyzed the effectiveness of the proposed Shapley value evaluation, as well as the influence of hyperparameters on task performance and search cost. We compare our Shapley-NAS with the state-of-the-art methods with respect to the accuracy, model complexity, and search cost. Finally, we empirically demonstrated the effectiveness of Monte-Carlo sampling estimation, as well as the high correlation between obtained Shapley value and task performance.

\subsection{Datasets and Implementation Details}
\textbf{CIFAR-10: } For the CNN search space on CIFAR-10, we employed the same operation space $\mathcal{O}$ as DARTS and set the initial channel number as 16. We utilized the partial connection strategy in PC-DARTS \cite{xu2019pc} to reduce memory overhead and increase batch size. We trained the supernet for 50 epochs (the first 15 epochs for warm-up) with a batch size of 256. The training set of CIFAR-10 was divided into two parts with equal size, one for optimizing network weights and the other for evaluating Shapley value. We set the number of samples $M$ to be 10 in the Monte-Carlo sampling and the early truncation threshold $\eta$ to be 0.5. The momentum coefficient $\mu$ and step size $\epsilon$ were assigned to 0.8 and 0.1 respectively. At the evaluation phase, We simply followed the DARTS experimental settings for fair comparison and retrained the network from scratch for 600 epochs.

\textbf{ImageNet: }ImageNet contains about 1.2 million training and 50K validation images from 1000 categories, which is much more challenging than CIFAR-10. We randomly sampled $10\%$ and $2.5\%$ images from the entire 1.3M training set of ImageNet for training network weights and estimating Shapley value respectively. The supernet was trained for 50 epochs with batch size 1024 and the architecture parameters remained frozen in the first 25 epochs. The other hyper-parameters were the same with CIFAR-10. At the evaluation stage, we trained the network from scratch for 250 epochs by an SGD optimizer with a linearly decayed learning rate initialized as 0.5, a momentum of 0.9, and a weight decay of $3\times 10^{-5}$.

\textbf{NAS-Bench-201: } NAS-Bench-201 is a popular benchmark to analyze NAS algorithms, as it provides performance of all candidate architectures which can be directly obtained by querying. In the search space of NAS-Bench-201, the operation set $\mathcal{O}$ has 5 elements and each cell contains 4 nodes, leading to a total search space of 15,625 architectures. NAS-Bench-201 supports three datasets, CIFAR-10, CIFAR-100, and ImageNet-16-120, and more details about the datasets can be found in their paper \cite{dong2020bench}. Specifically, we acquired the task-specific performance by directly searching on the evaluation dataset, and obtained the mean and standard deviation for the best architecture from 4 independent runs with different random seeds.
\begin{table}[t]
 \caption{\small{The test error(\%) of different search algorithms on S1-S4. DARTS+Shapley denotes the combination of DARTS and Shapley value evaluation, and $*$ means freezing $\alpha$ during the search. }}
 \vspace{-0.5em}
\label{tab:abl1}

 \renewcommand\arraystretch{1.0}
 \centering
 \resizebox{0.42\textwidth}{!}{
 
\begin{tabular}{c|c|c|c|c|c}
\hline   
\textbf{Dataset} &\textbf{Method} &\textbf{S1} & \textbf{S2} & \textbf{S3} & \textbf{S4}   \\\hline
             \multirow{4}*{C10} & DARTS & 3.84 & 4.85 & 3.34 & 7.20  \\
             & DARTS+Shapley & 3.11 & 2.92 &2.58 & 3.45  \\
            & DARTS+Shapley$^{*}$ & 2.95 & 2.84& 2.67 & 2.94 \\
            & Shapley-NAS & 2.82 & 2.55 & 2.42 & 2.63  \\ \hline             
            \multirow{4}*{C100} 
            & DARTS         & 29.46 & 26.05 & 28.90& 22.85 \\
             & DARTS+Shapley & 28.21 & 24.51 & 23.67& 22.78  \\
      & DARTS+Shapley$^{*}$ & 25.24& 24.66& 22.39 &  22.15\\
            & Shapley-NAS &23.60 & 22.77 & 21.92& 21.53  \\ \hline 
            
             \multirow{4}*{SVHN} 
             & DARTS         & 4.58 & 3.53 &3.41 & 3.05  \\
             & DARTS+Shapley & 2.59 & 2.72 & 2.83 & 2.65  \\
            & DARTS+Shapley$^{*}$ & 2.88& 2.64 & 2.49 & 2.58 \\
            & Shapley-NAS     & 2.36 & 2.43 & 2.34 & 2.41  \\ \hline    
\end{tabular}}
\vspace{-1.5em}
\end{table}

\subsection{Ablation Study}
 \textbf{Effectiveness of Shapley value evaluation: }To verify the effectiveness of Shapley-NAS, we conducted experiments on 4 simplified search spaces S1-S4 proposed by \cite{zela2019understanding} on CIFAR-10, CIFAR100, and SVHN. We first combined the proposed Shapley value evaluation method with DARTS (denoted as DARTS+Shapley in \cref{tab:abl1}, by only applying Shapley value evaluation at the final discretization step, $\ie$ selecting operations based on their Shapley values instead of $\alpha$. Moreover, we also tested the performance under the same setting but keeping $\alpha$ frozen, denoted as DARTS+Shapley$^{*}$. As shown in \cref{tab:abl1}, DARTS achieves competitive results with the proposed Shapley evaluation method, even when $\alpha$ is not optimized in the training. Notably, our Shapley-NAS still outperforms DARTS+Shapley and DARTS+Shapley$^{*}$, since taking Shapley value into the supernet optimization can further alleviate the problem caused by gradient-based NAS methods.

\begin{figure}

  \begin{subfigure}{0.48\linewidth}
  \includegraphics[ width=1.0\textwidth]{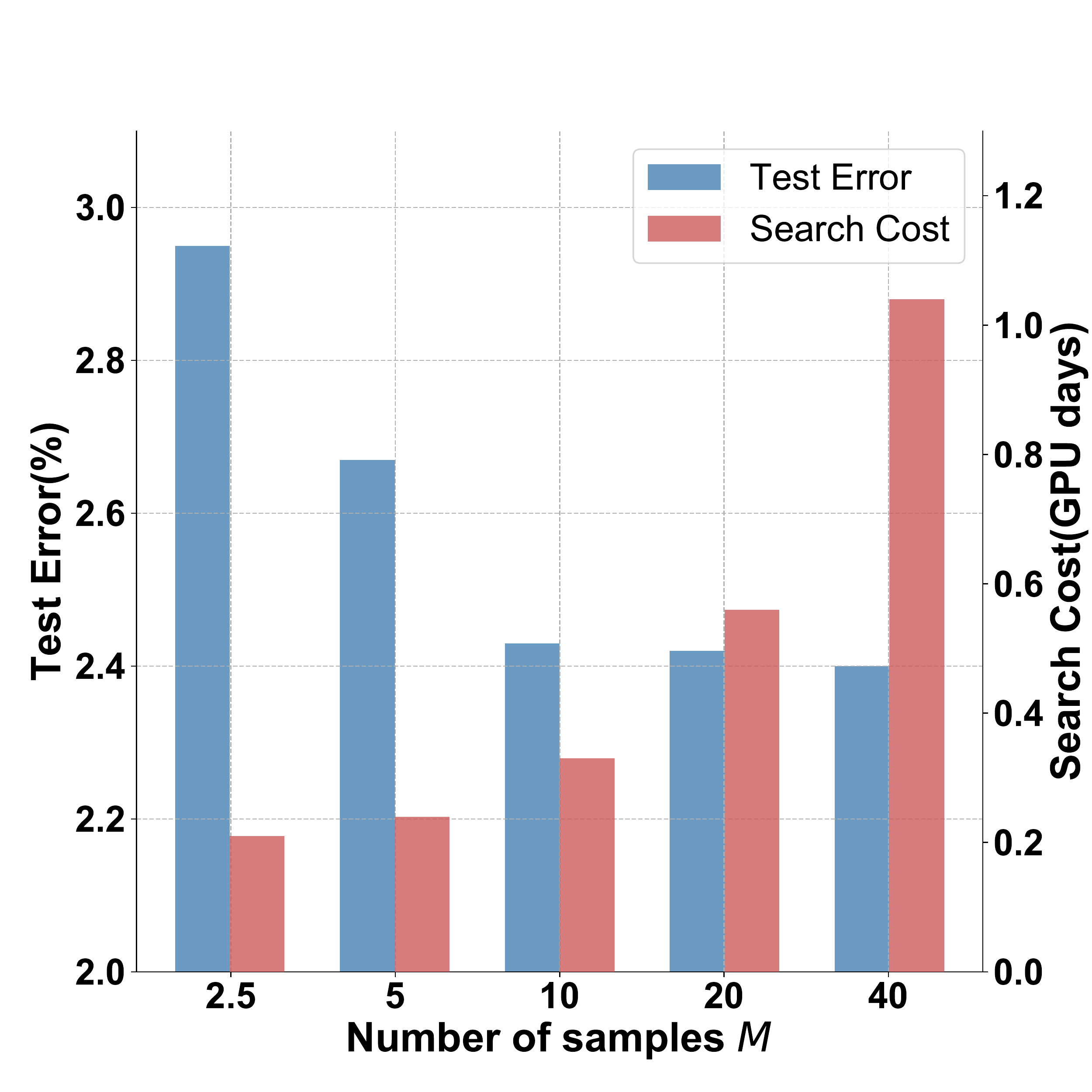}
    
    \caption{\scriptsize{Varying number of samples $M$}}
    \label{fig:short-a}
  \end{subfigure}
   \hfill
  \begin{subfigure}{0.48\linewidth}
    \includegraphics[ width=1.0\textwidth]{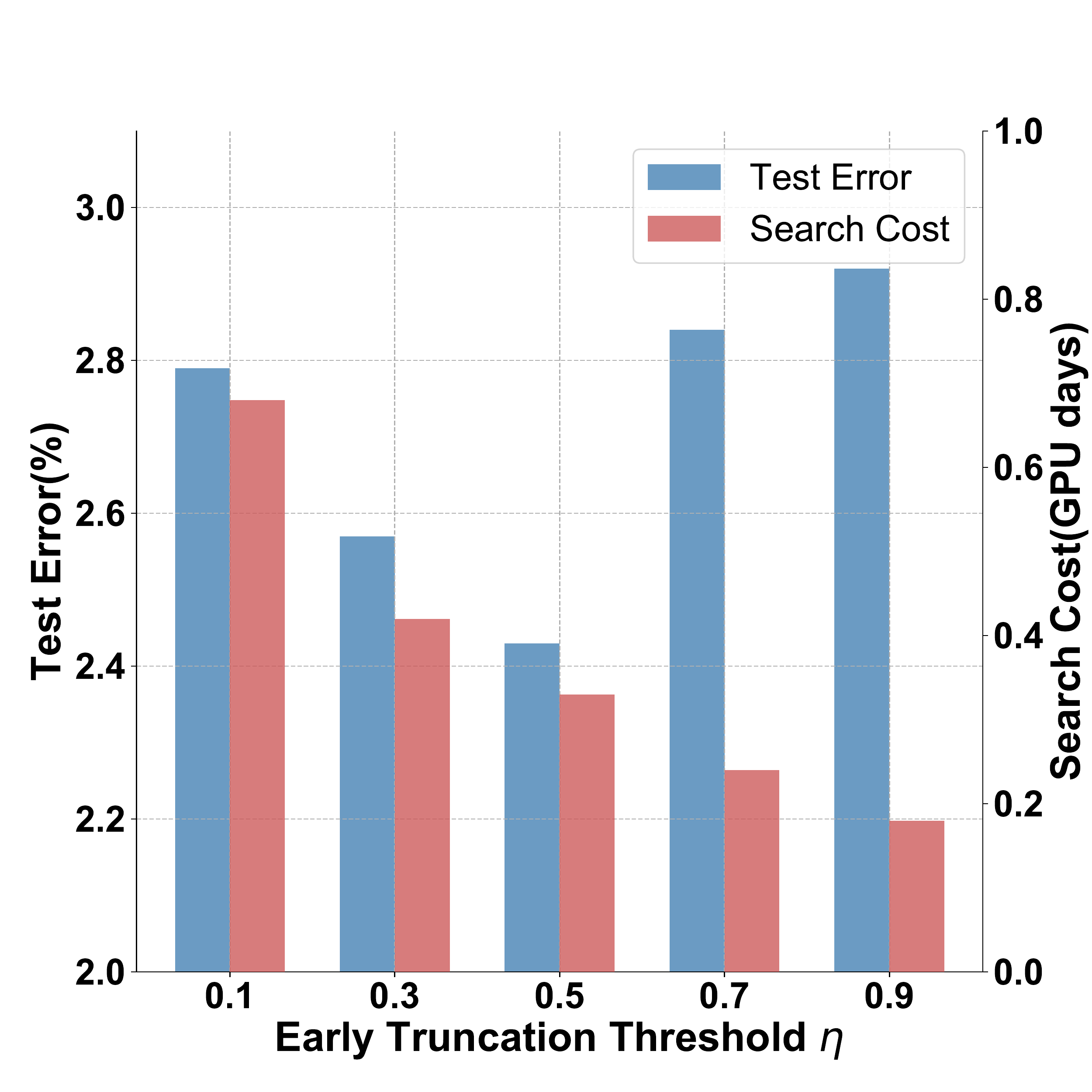}
    
    \caption{\scriptsize{Varying early truncation threshold $\eta$}}
    \label{fig:short-b}
  \end{subfigure}
  \vspace{-0.5em}
    \caption{\small{ The test error (\%) and search cost (GPU days) of the proposed method on CIFAR-10  with (a) different number of samples and (b) various thresholds of early truncation in the Monte-Carlo sampling for Shapley value estimation.}}
  \label{fig:abl3}
  \vspace{-1.5em}
\end{figure}

\textbf{Influence of samples times $M$ and early truncation threshold $\eta$: }We also explored the influence of sampling times $M$ and early truncation threshold $\eta$ in the Monte-Carlo sampling algorithm. The values of sampling times $M$ and early truncation threshold $\eta$ are significant for accurate Shapley value estimation, which also affect the overall search cost. \cref{fig:abl3} shows the test error (\%) and search cost (GPU days) on CIFAR-10 with various $M$ and $\eta$. Reducing the number of samples results in lower search cost while degrading the performance since the sampling is not enough to make an accurate estimation. However, the estimation accuracy with samples larger than 10 is not sensitive to the number of samples, and we choose $M=10$ for search efficiency. Meanwhile, medium $\eta$ also achieves the best accuracy-complexity trade-off as it mitigates the fluctuation of sampling and reduces the search cost.

  \begin{table*}[htbp]
    \centering
    \caption{\small{The test error (\%) and parameter storage cost (M) of the final architectures w.r.t. different values of momentum coefficient $\mu$ and different assignments of step size $\epsilon$. }}
    \vspace{-0.5em}
    \resizebox{0.95\textwidth}{!}{
    \begin{threeparttable}{
\begin{tabular}{c|cc|cc|cc|cc}
\hline
     
\multirow{2}*{\textbf{step size $\epsilon$}} & \multicolumn{2}{c|}{\textbf{$\mu=0.2$}}& \multicolumn{2}{c|}{\textbf{$\mu=0.5$}} & \multicolumn{2}{c|}{\textbf{$\mu=0.8$}}& \multicolumn{2}{c}{\textbf{$\mu=0.9$}}
  \\ \cline{2-9}
  &\textbf{Test Error(\%)} & \textbf{Params(M)}  &\textbf{Test Error(\%)} & \textbf{Params(M)}
  &\textbf{Test Error(\%)} & \textbf{Params(M)} &\textbf{Test Error(\%)} & \textbf{Params(M)}  \\\hline
             0.01 &$2.89\pm 0.21 $&$4.0 $&$2.87\pm 0.16$ &$3.7 $
             &$2.67\pm 0.06 $ &$3.5 $ &$2.74\pm 0.11 $ &$3.8 $   \\
             0.05 &$2.85\pm 0.18 $&$3.6 $&$2.79\pm 0.12$ &$3.4 $
             &$2.55\pm 0.07 $ &$3.2 $ &$2.68\pm 0.07 $ &$3.5 $   \\
            0.1 &$2.82\pm 0.11 $&$3.7 $&$2.66\pm 0.10$ &$3.3 $
             &$2.47\pm 0.04 $ &$3.4 $ &$2.61\pm 0.06 $ &$4.1 $   \\
            0.5 &$2.92\pm 0.19 $&$3.5 $&$2.84\pm 0.13$ &$4.2 $
             &$2.71\pm 0.12 $ &$3.8 $ &$2.83\pm 0.15 $ &$3.9 $   \\ \hline

\end{tabular}
	 }
\end{threeparttable}
	}
       
    \label{tab:abl2}
    \vspace{-1.2em}
\end{table*}

\textbf{Impact of momentum coefficient $\mu$ and step size $\epsilon$: }To investigate the influence of momentum coefficient $\mu$ and step size $\epsilon$ on test accuracy, we implemented the architecture parameter assignment with different $\mu$ and $\epsilon$. The test error range and model parameter cost are demonstrated in \cref{tab:abl2}, where medium $\epsilon$ outperforms other values. Small step sizes fail to achieve the optimal distribution when reaching the maximum update iterations, and large step sizes make the supernet optimization hard to converge. With the increase of $\mu$, the training stabilization becomes enforced, where $\mu$ with $0.8$ achieves the best accuracy. 

 \begin{table}[t]
    \centering
    \caption{Comparison with state-of-the-art image classifiers on CIFAR-10. Means and standard deviations of our Shapley-NAS are obtained by repeated experiments with 4 random seeds.}
    \vspace{-0.5em}
    \resizebox{.43\textwidth}{!}{
    \begin{threeparttable}{
\begin{tabular}{lccc}
\hline
            \textbf{Architecture} & \textbf{\tabincell{c}{Test Error\\(\%)}} & \textbf{\tabincell{c}{Params\\(M)}} & \textbf{\tabincell{c}{Search Cost\\(GPU days)}}  \\ \hline
 DenseNet-BC \cite{huang2017densely} & 3.46 & 25.6 & -  \\ \hline
 NASNet-A \cite{zoph2018learning} & 2.65 & 3.3 & 2000  \\
            AmoebaNet-A \cite{real2019regularized} & $3.34\pm 0.06$ & 3.2 & 3150  \\
            AmoebaNet-B \cite{real2019regularized} & $2.55\pm0.05$ & 2.8 & 3150  \\
            PNAS \cite{liu2018progressive} & $3.41\pm 0.09$ & 3.2 & 225  \\
            ENAS \cite{pham2018efficient} & 2.89 & 4.6 & 0.5 \\
            NAONet \cite{luo2018neural} & 3.53 & 3.1 & 0.4  \\
            RandomNAS \cite{li2020random} & $2.85\pm 0.08$ & 4.3 & 2.7  \\ \hline
            DARTS (1st order) \cite{liu2018darts} & $3.00\pm 0.14$ & 3.3 & 0.4  \\
            DARTS (2nd order) \cite{liu2018darts} & $2.76\pm 0.09$ & 3.3 & 1.0  \\ 
            SNAS(moderate)  \cite{xie2018snas} & $2.85\pm 0.02$ & 2.8 & 1.5  \\
            GDAS \cite{dong2019searching} & 2.93 & 3.4 & 0.3  \\
            BayesNAS \cite{zhou2019bayesnas} & $2.81\pm 0.04$ & 3.4 & 0.2  \\
            ProxylessNAS \cite{cai2018proxylessnas} & 2.08 & 5.7 & 4.0  \\
            
            P-DARTS \cite{chen2019progressive} & 2.50 & 3.4 & 0.3  \\ 
            PC-DARTS \cite{xu2019pc} & $2.57\pm 0.07$ & 3.6 & 0.1  \\ 
            SGAS (Cri 1. avg) \cite{li2020sgas} & $2.66\pm 0.24$ & 3.7 & 0.25  \\
            SDARTS-RS \cite{chen2020stabilizing} & $2.61\pm 0.02$ & 3.4 & 0.4  \\
            DrNAS \cite{chen2020drnas} & $2.54\pm 0.03$ & 4.0 & 0.4  \\
            DARTS+PT \cite{wang2021rethinking} & $2.61\pm 0.08$ & 3.0 & 0.8  \\
             \hline
            Shapley-NAS(avg.)  & $2.47\pm 0.04$ & 3.4 & 0.3 \\ 
            Shapley-NAS(best)  & $2.43$ & 3.6 & 0.3  \\ \hline

\end{tabular}
	 }
\end{threeparttable}
	}
       
    \label{cifar10}
    \vspace{-0.8em}
\end{table}

\subsection{Comparison with the State-of-the-art NAS Methods}
\cref{cifar10} shows the performance of Shapley-NAS on CIFAR-10 compared with the state-of-the-art NAS methods. Our Shapley-NAS achieves an average test error of $2.47\%$ while only using 0.3 GPU days, significantly surpassing the DARTS baseline in both search cost and accuracy. The test error of the best single run in our experiments is $2.43\%$, ranking top amongst popular NAS methods. Although ProxylessNAS \cite{cai2018proxylessnas} achieves a lower test error of $2.08\%$, it performs architecture search on a different space with heavy search cost. The low variance of the experimental results also demonstrates the stability of the proposed search method.
\begin{table}[htbp]
    \centering
    \caption{Comparison with state-of-the-art image classifiers on ImageNet under the mobile setting \cite{liu2018darts}. $\dagger$ indicates the results obtained by searching on ImageNet, otherwise on CIFAR-10.}
    \vspace{-0.5em}
    \resizebox{0.478\textwidth}{!}{
    \begin{threeparttable}
    \begin{tabular}{lcccc}
    \hline
    
\textbf{Architecture} & \textbf{\tabincell{c}{Test Error\\(\%)}} & \textbf{\tabincell{c}{Params\\(M)}} & \textbf{\tabincell{c}{$\times +$\\(M)}}& \textbf{\tabincell{c}{Search Cost\\(GPU days)}}  \\ \hline

    Inception-v1 \cite{szegedy2015going} & 30.1  & 6.6 & 1448 & -  \\
    MobileNet \cite{howard2017mobilenets} & 29.4 & 4.2 & 569 & -  \\
    ShuffleNet $2\times$ (v1) \cite{zhang2018shufflenet} & 26.4 &  $\sim 5$ & 524 & -  \\
    ShuffleNet $2\times$ (v2) \cite{ma2018shufflenet} & 25.1 &  $\sim 5$ & 591 & -  \\ \hline
    
    NASNet-A \cite{zoph2018learning} & 26.0 & 5.3 & 564 & 2000  \\
    AmoebaNet-C \cite{real2019regularized} & 24.3 & 6.4 & 570 & 3150 \\
    PNAS \cite{liu2018progressive} & 25.8 & 5.1 & 588 & 225  \\
    MnasNet-92 \cite{tan2019mnasnet} & 25.2 & 4.4 & 388 & - \\ \hline
    
    DARTS (2nd) \cite{liu2018darts} & 26.7 & 4.7 & 574 & 1.0  \\
    SNAS (mild) \cite{xie2018snas} & 27.3 & 4.3 & 522 & 1.5  \\
    GDAS \cite{dong2019searching} & 26.0 & 5.3 & 545 & 0.3  \\
    BayesNAS \cite{zhou2019bayesnas} & 26.5 & 3.9 & - & 0.2  \\
    ProxylessNAS (GPU) \cite{cai2018proxylessnas}\tnote{$\dagger$} & 24.9 & 7.1 & 465 & 8.3  \\
    P-DARTS \cite{chen2019progressive} & 24.4 & 4.9 & 557 & 0.3  \\  
    PC-DARTS \cite{xu2019pc} & 25.1 & 5.3 & 586 & 0.1  \\
    PC-DARTS \cite{xu2019pc}\tnote{$\dagger$} & 24.2 &  5.3 & 582 &3.8 \\  
    SGAS (Cri 1. best) \cite{li2020sgas} & 24.2 &  5.3 & 585 & 0.25  \\ 
    SDARTS-ADV \cite{chen2020stabilizing} & 25.6 & 6.1 & - & 0.4  \\
    DrNAS \cite{chen2020drnas}\tnote{$\dagger$} & 24.2  & 5.2 & 644 & 3.9  \\
    DARTS+PT \cite{wang2021rethinking}\tnote{$\dagger$} & 25.5  & 4.7 & 538 & 3.4  \\ \hline
    Shapley-NAS & 24.3  & 5.1& 566 & 0.3  \\
    Shapley-NAS\tnote{$\dagger$} & 23.9  & 5.4 & 582 & 4.2  \\ \hline
    \end{tabular}
    \end{threeparttable}}
    \label{imagenet_table}
    \vspace{-1.5em}
\end{table}

\begin{table*}[t]

    \centering
    \caption{Comparison results with state-of-the-art NAS methods on NAS-Bench-201. $\dagger$ denotes the results are obtained by searching on CIFAR-10, otherwise by directly searching on the evaluation dataset.}
    \vspace{-0.5em}
    \resizebox{0.9\textwidth}{!}{
    \begin{threeparttable}
    \begin{tabular}{lccccccc}
    \hline
    \multirow{2}*{\textbf{Method}} & \multicolumn{2}{c}{\textbf{CIFAR-10}} & \multicolumn{2}{c}{\textbf{CIFAR-100}} & \multicolumn{2}{c}{\textbf{ImageNet-16-120}} \\ \cline{2-7}
    & \textbf{validation} & \textbf{test} & \textbf{validation} & \textbf{test} & \textbf{validation} & \textbf{test} \\ \hline
    ResNet \cite{he2016deep} & 90.83 & 93.97 & 70.42 & 70.86 & 44.53 & 43.63 \\ \hline
    Random (baseline) & $90.93\pm 0.36$ & $93.70\pm 0.36$ & $70.60\pm 1.37$ & $70.65\pm 1.38$ & $42.92\pm 2.00$ & $42.96\pm 2.15$ \\
    RSPS \cite{li2020random} & $84.16\pm 1.69$ & $87.66\pm 1.69$ & $45.78\pm 6.33$ & $46.60\pm 6.57$ & $31.09\pm 5.65$ & $30.78\pm 6.12$ \\
    REINFORCE \cite{zoph2018learning}\tnote{$\dagger$} & $91.09\pm 0.37$ & $93.85\pm 0.37$ & $71.61\pm 1.12$ & $71.71\pm 1.09$ & $45.05\pm 1.02$ & $45.24\pm 1.18$ \\     
    ENAS \cite{pham2018efficient} & $39.77\pm 0.00$ & $54.30\pm 0.00$ & $10.23\pm 0.12$ & $10.62\pm 0.27$ & $16.43\pm 0.00$ & $16.32\pm 0.00$ \\ \hline
    DARTS \cite{liu2018darts}\tnote{$\dagger$} & $39.77\pm 0.00$ & $54.30\pm 0.00$ & $15.03\pm 0.00$ & $15.61\pm 0.00$ & $16.43\pm 0.00$ & $16.32\pm 0.00$ \\
    DARTS \cite{liu2018darts} & $39.77\pm 0.00$ & $54.30\pm 0.00$ & $38.57\pm 0.00$ & $38.97\pm 0.00$ & $18.87\pm 0.00$ & $18.41\pm 0.00$ \\
    SNAS \cite{xie2018snas} & $90.10\pm 1.04$ & $92.77\pm 0.83$ & $69.69\pm 2.39$ & $69.34\pm 1.98$ & $42.84\pm 1.79$ & $43.16\pm 2.64$ \\
    GDAS \cite{dong2019searching} & $90.01\pm 0.46$ & $93.23\pm 0.23$ & $24.05\pm 8.12$ & $24.20\pm 8.08$ & $40.66\pm 0.00$ & $41.02\pm 0.00$ \\
    PC-DARTS \cite{xu2019pc} & $89.96\pm 0.15$ & $93.41\pm 0.30$ & $67.12\pm 0.39$ & $67.48\pm 0.89$ & $40.83\pm 0.08$ & $41.31\pm 0.22$ \\
    iDARTS \cite{zhang2021idarts}\tnote{$\dagger$} & $89.96\pm 0.60$ & $93.58\pm 0.32$ & $70.57\pm 0.24$ & $70.83\pm 0.48$ & $40.38\pm 0.59$ & $40.89\pm 0.68$ \\
    DrNAS \cite{chen2020drnas} & ${91.55\pm 0.00}$ & ${94.36\pm 0.00}$ & ${73.49\pm 0.00}$ & ${73.51\pm 0.00}$ & ${46.37\pm 0.00}$ & ${46.34\pm 0.00}$ \\ \hline
    
    \textbf{Shapley-NAS} & $\bf{91.61\pm 0.00}$ & $\bf{94.37\pm 0.00}$ & $\bf{73.49\pm 0.00}$ & $\bf{73.51\pm 0.00}$ & $\bf{46.57\pm 0.08}$ & $\bf{46.85\pm 0.12}$ \\ \hline
    \textbf{optimal} & 91.61 & 94.37 & 73.49 & 73.51 & 46.77 & 47.31 \\ \hline
    \end{tabular}

	\end{threeparttable}
	}
       
    \label{nasbench201_table}
    \vspace{-0.5mm}
\end{table*}

The comparison results on ImageNet with other methods are demonstrated in \cref{imagenet_table}. We follow the mobile setting in \cite{liu2018darts} for ImageNet, where the number of multiply-add operations (“$\times +$") is restricted to be less than 600M. We trained the best-found architecture on CIFAR-10 to evaluate its transferability to ImageNet and obtained a competitive result with $24.3\% / 7.3\%$ top-1/5 test error, verifying the generalization ability of our Shapley-NAS. We also evaluated the optimal architecture directly searched on ImageNet and obtain a top-1/5 test error of $23.9\% / 7.2\%$, which outperforms all other NAS methods with light search cost. Notably, despite the outstanding performance of DrNAS, it has a number of multiply-add operations much over 600M. By contrast, out Shapley-NAS never violates the mobile setting while achieving competitive results.

For NAS-Bench-201, our Shapley-NAS achieves outstanding performance with $94.37\%$, $73.51\%$, and $46.85\%$ test accuracy on CIFAR-10, CIFAR-100, and ImageNet-16-120 respectively, as shown in \cref{nasbench201_table}. Notably, we obtain the global optimal architectures on CIFAR-10 and CIFAR-100, which indicates that the proposed method can identify important operations and derive the best architecture from the large search space. On the ImageNet-16-120 dataset, we also acquire a near-optimal solution, which outperforms the state-of-the-art algorithms, again verifying the effectiveness of our Shapley-NAS.

\begin{figure}
   \centering
  \begin{subfigure}{0.48\linewidth}
  \includegraphics[ width=1.0\textwidth]{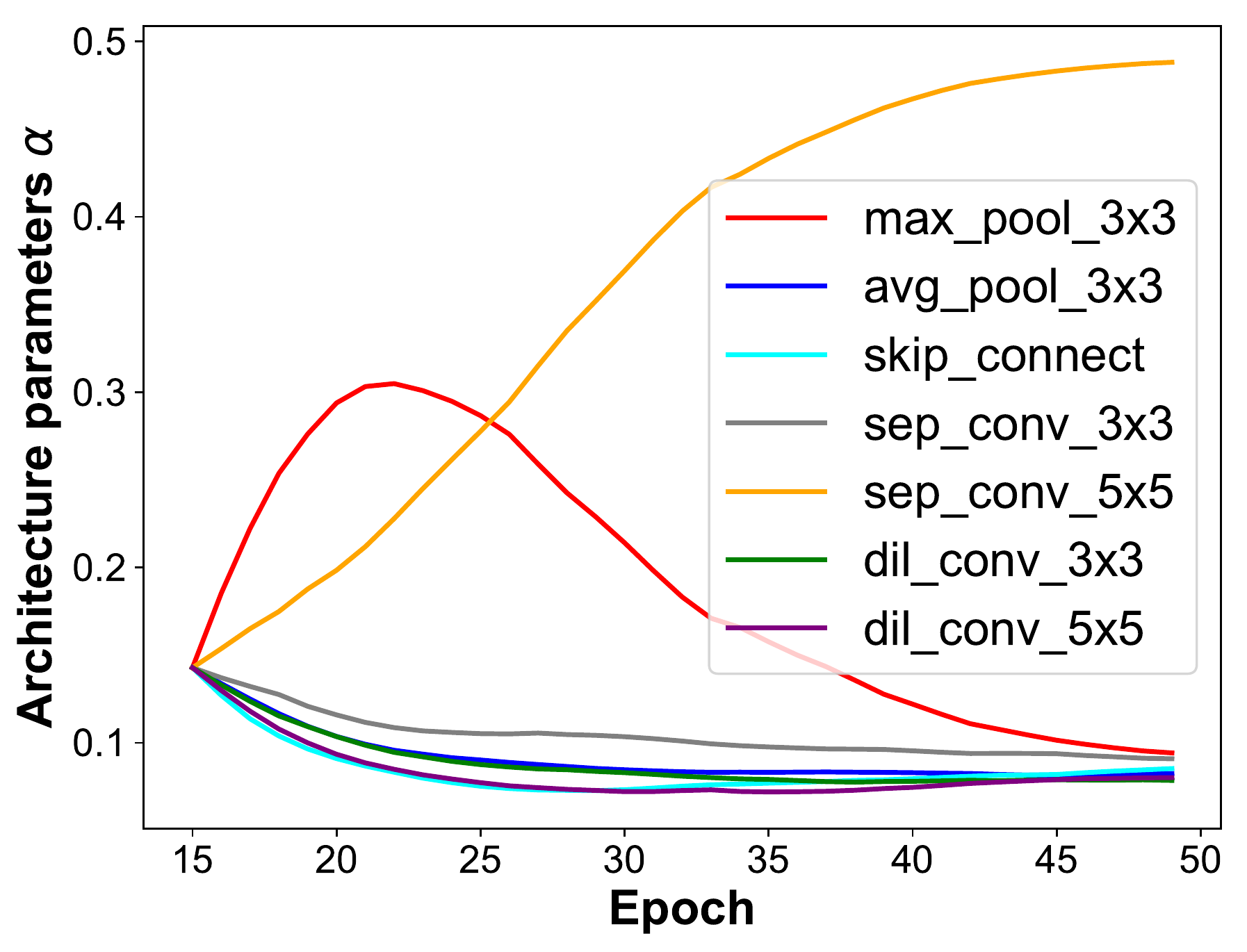}
    \caption{{Normal cell}}
    \label{fig:converge-1}
  \end{subfigure}
  \begin{subfigure}{0.48\linewidth}
    \includegraphics[ width=1.0\textwidth]{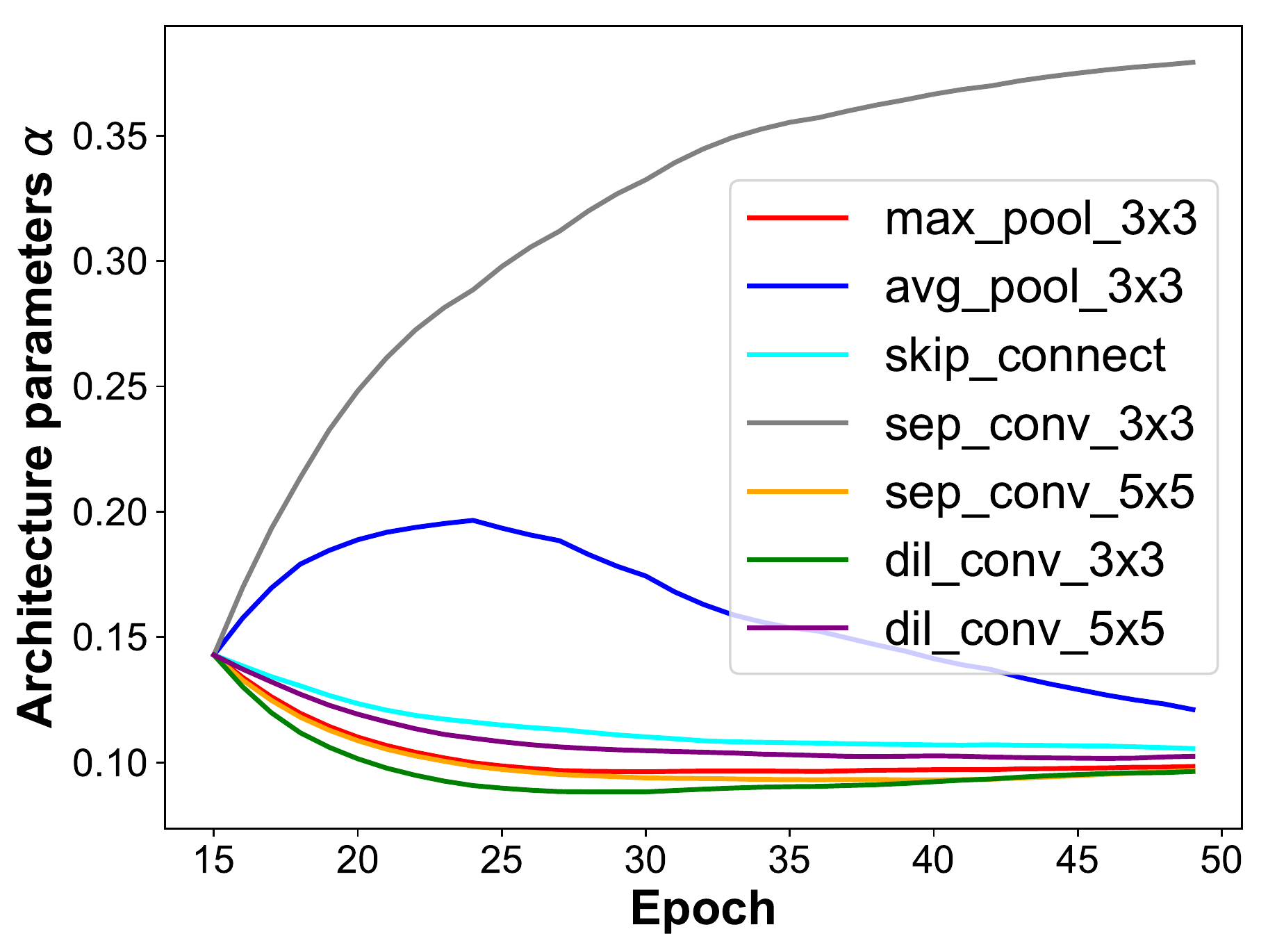}
    \caption{{Reduction cell}}
    \label{fig:converge-2}
  \end{subfigure}
  \vspace{-0.5em}
    \caption{{ The evolution of architecture parameters $\alpha$ by estimating Shapley value based on Monte-Carlo Sampling. (a) The curves of $\alpha$ on the first edge of the normal cell. (b) The curves of $\alpha$ on the first edge of the reduction cell.}}
  \label{fig:evo}
 \vspace{-1.5em}
\end{figure}
\begin{figure*}
   \centering
  \begin{subfigure}{0.3\linewidth}
  \includegraphics[ width=1.0\textwidth]{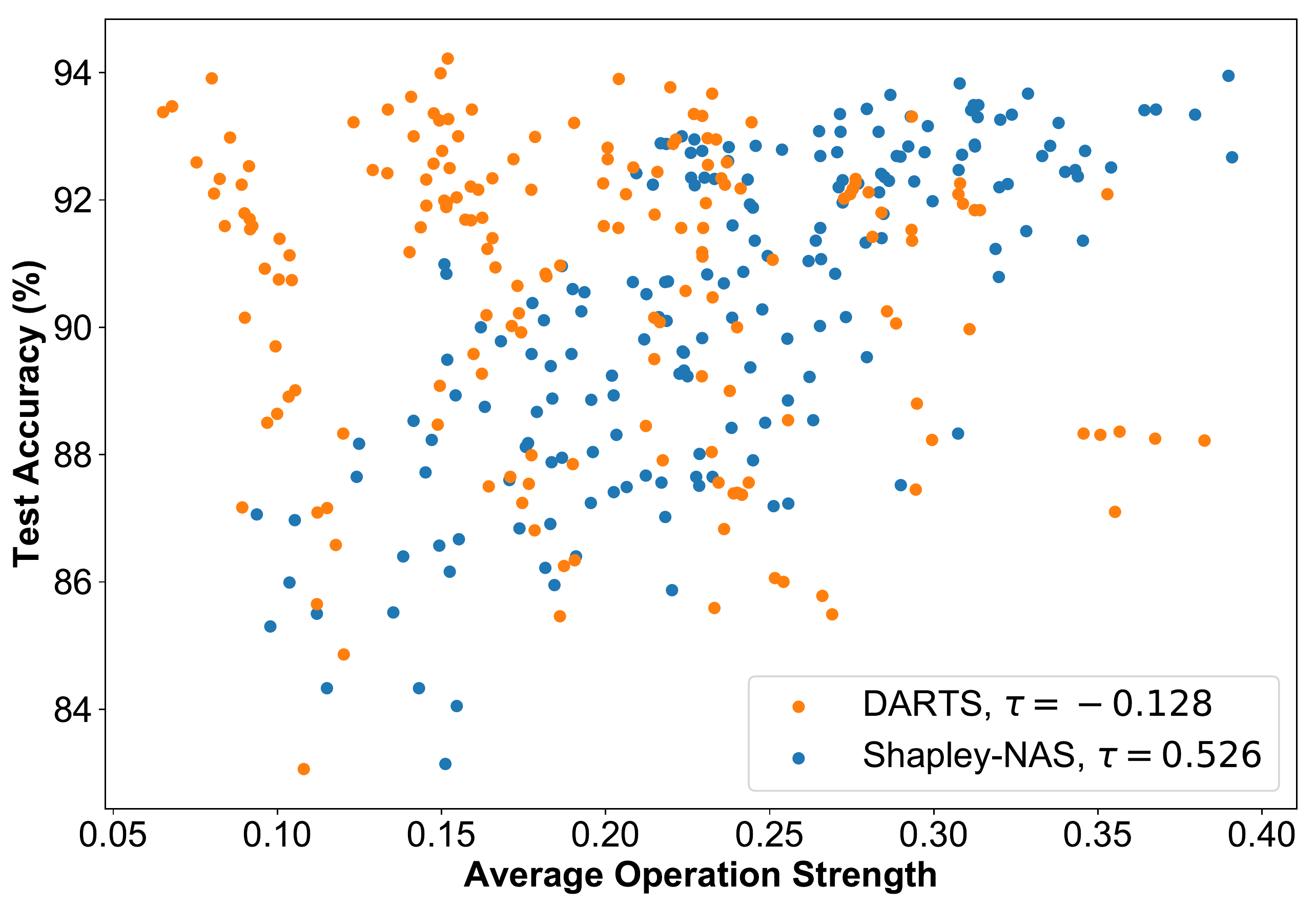}
    \caption{{CIFAR-10}}
    \label{fig:cifar10}
  \end{subfigure}
   \hfill 
  \begin{subfigure}{0.3\linewidth}
    \includegraphics[ width=1.0\textwidth]{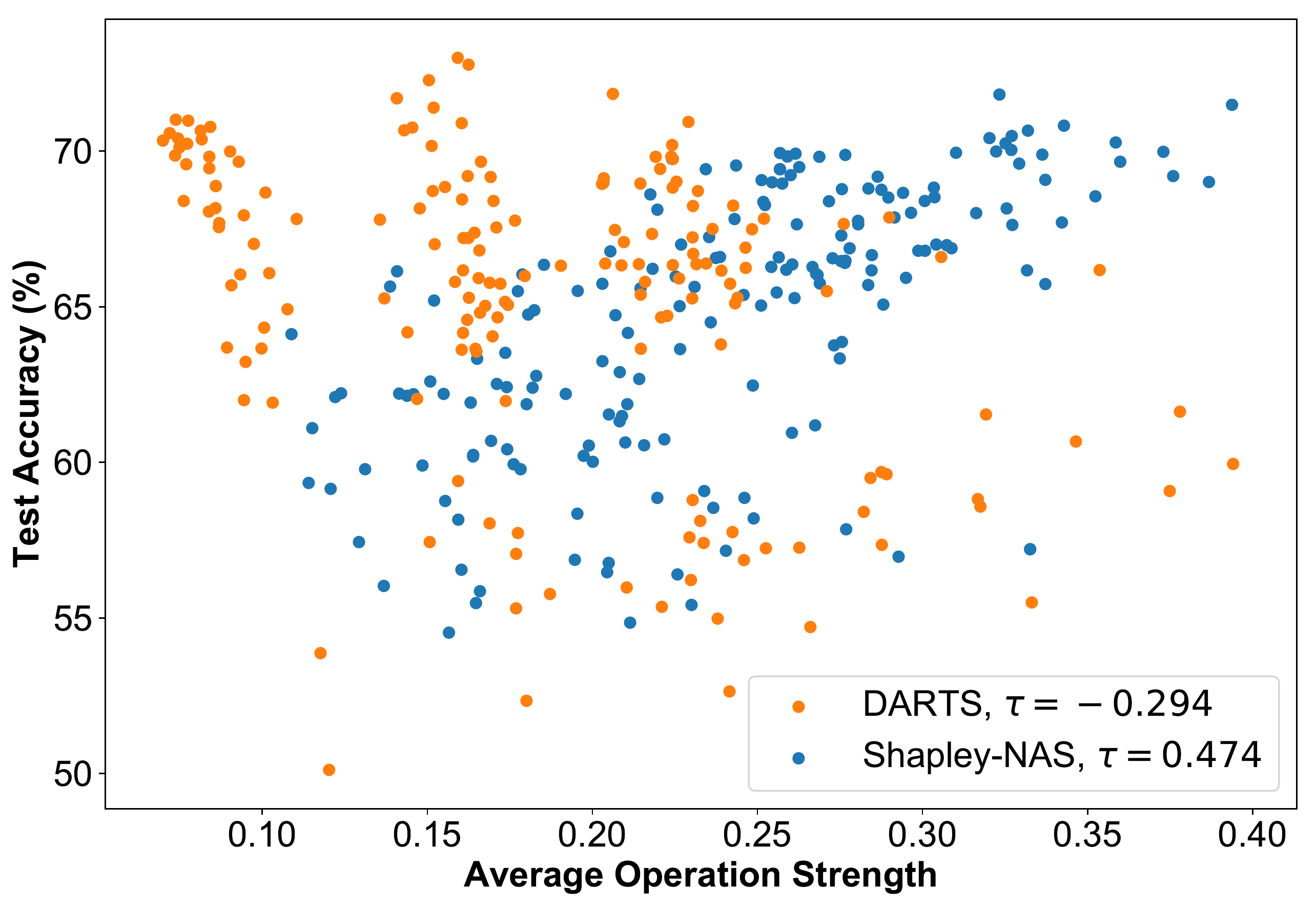}
    \caption{{CIFAR-100}}
    \label{fig:cifar100}
  \end{subfigure}
 	\hfill
    \begin{subfigure}{0.3\linewidth}
    \includegraphics[ width=1.0\textwidth]{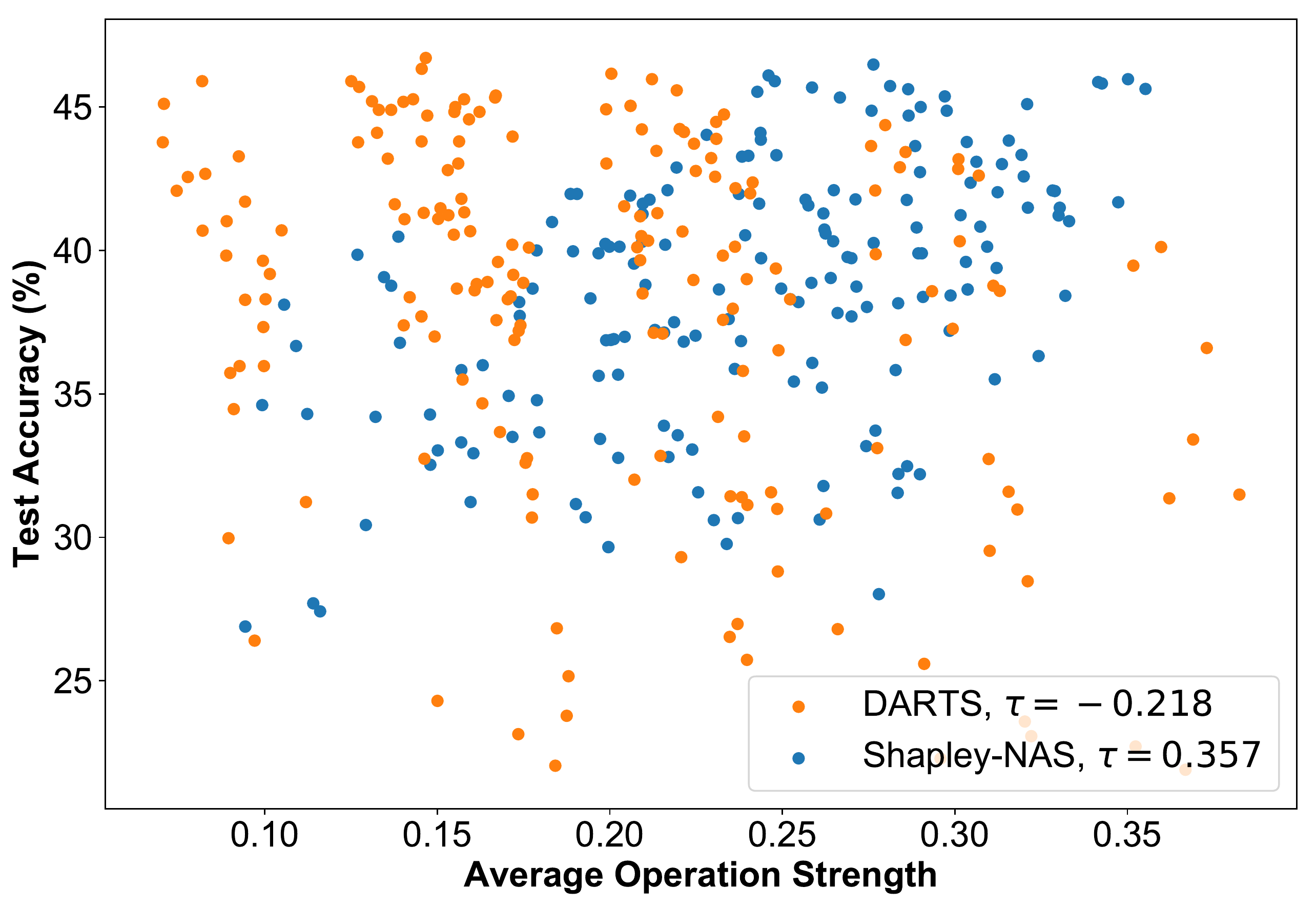}
    \caption{{ImageNet-16-120}}
    \label{fig:imagenet}
  \end{subfigure}
  \vspace{-0.5em}
    \caption{{Correlation between operation strength and test accuracy of 200 sampled architectures using DARTS and our Shapley NAS on NAS-Bench-201. The average operation strength is obtained by the magnitude of corresponding architecture parameters in the supernet, and $\tau$ is the Kendall Tau coefficient that measures the correlation.}}
  \label{fig:corr}
  \vspace{-1.5em}
\end{figure*}

 \section{Performance Analysis}
\textbf{Shapley value estimation by Monte-Carlo Sampling: }To verify the effectiveness of Monte-Carlo Sampling for Shapley value approximation, we plot the architecture parameters evolution on the first edge of normal and reduction cells in \cref{fig:evo}. Note that the curves of the first 15 epochs for warm-up are not presented. As \cref{fig:converge-1} shows, although the \emph{max\_pool\_3x3} operation is larger than all other operations at the start, the operation \emph{sep\_conv\_5x5} finally becomes the strongest operation since it has the most contribution to the supernet along with the training process. While in \cref{fig:converge-2}, the operation \emph{sep\_conv\_3x3} becomes dominant after several epochs, while other operations gradually converge to be very weak. The supernet gradually converges to the final derived architecture using the proposed estimation. Moreover, the architecture parameters are differentiated to make the $\arg \max$ selection more reliable.

\textbf{Correlation between Shapley value and task performance:  }We investigate the correlation between Shapley value of operations and real task performance on NAS-Bench-201. After the search phase, we sample 200 discrete architectures from the search space and compute their corresponding operation strength by averaging the magnitude of architecture parameters. Then we plot the test accuracy obtained by directly querying along with the computed operation strength of DARTS and our Shapley-NAS. We use the Kendall Tau coefficient to measure the correlation, and the results on CIFAR-10, CIFAR-100, and ImageNet-16-120 are shown in \cref{fig:corr}. The Shapley value of operations has a higher correlation with the test accuracy ($\tau = 0.526, 0.474, 0.357$ respectively), while the magnitude of $\alpha$ in DARTS is almost entirely uncorrelated with the final task performance. It indicates the effectiveness of Shapley value to help us discover optimal architectures with superior performance during the evaluation phase.

\section{Conclusion and Discussion}
In this paper, we have presented Shapley-NAS, a Shapley value based operation contribution evaluation method for neural architecture search. Since the architecture parameters updated by gradient descent in DARTS cannot reveal the actual operation importance in general, we propose to directly evaluate the marginal contribution of operations on accuracy via Shapley value. Specifically, the Shapley value of operations can be efficiently approximated by Monte-Carlo sampling based algorithm with early truncation, thus enabling the optimization of the supernet whose architecture parameters are directly updated with the operation contribution. Shapley-NAS achieves state-of-the-art performance on CIFAR-10, ImageNet, and NAS-Bench-201 benchmarks, which proves its effectiveness to identify the optimal architectures with the most important operations in neural architecture search.

\vspace{0.5em}
\textbf{Limitations:  }The exact computation of Shapley value is expensive on common search space since it needs exponential times of evaluations, and the resulting heavy search burdens would limit the practical applications for task-specific network deployment. Therefore, to evaluate operation contribution during the architecture search, we utilize the Monte-Carlo sampling method which gives an unbiased approximation of Shapley value. Despite being computationally efficient, it might not be as accurate as the exact computation via enumerating all possible subsets.   
\vspace{-0.5em}
\section*{Acknowledgements}
This work was supported in part by the National Key Research and Development Program of China under Grant 2017YFA0700802, in part by the National Natural Science Foundation of China under Grant 62125603 and Grant U1813218, in part by a grant from the Beijing Academy of Artificial Intelligence (BAAI), and in part by a grant from the Institute for Guo Qiang, Tsinghua University.

{\small
\bibliographystyle{ieee_fullname}
\bibliography{egbib}
}

\end{document}